\documentclass[journal]{IEEEtran}
\IEEEoverridecommandlockouts
\usepackage{cite}
\usepackage{booktabs}
\usepackage{amsmath,amssymb,amsfonts}
\usepackage{algorithmic}
\usepackage{graphicx}
\usepackage{textcomp}
\usepackage{algorithmic}
\usepackage{array}
\usepackage{tabu}
\usepackage{tabularx}
\usepackage{svg}
\newcommand{\eatme}[1]{ }
\usepackage[inline]{enumitem}
\usepackage{float}
\usepackage{cite}
\usepackage{algorithmic}
\usepackage{array}
\usepackage{tabu}
\usepackage{tabularx}
\usepackage{graphicx}
\usepackage{textcomp}
\usepackage{subcaption,siunitx,booktabs}
\usepackage{multirow}
\usepackage{booktabs} 
\usepackage{ragged2e}
\usepackage{tabularx,tabulary}
\usepackage{adjustbox}
\usepackage{caption}
\renewcommand{\thetable}{\arabic{table}}   
\renewcommand{\thefigure}{\arabic{figure}}
\usepackage{xcolor}

\usepackage{makecell}
\usepackage{fixltx2e}

\def\BibTeX{{\rm B\kern-.05em{\sc i\kern-.025em b}\kern-.08em
    T\kern-.1667em\lower.7ex\hbox{E}\kern-.125emX}}

\begin{document}

\title{\vspace{0.25cm} TGDT: A Temporal Graph-based Digital Twin for Urban Traffic Corridors\\
}

\makeatletter
\newcommand{\linebreakand}{%
  \end{@IEEEauthorhalign}
  \hfill\mbox{}\par
  \mbox{}\hfill\begin{@IEEEauthorhalign}
}


\author{Nooshin Yousefzadeh$^{1}$, Rahul Sengupta$^{1}$, Jeremy Dilmore$^{2}$, and Sanjay Ranka$^{1}$\\
\textit{$^{1}$ University of Florida, Gainesville, FL, USA} \\
\textit{$^{2}$ Florida Department of Transportation} \\
\textit{nooshinyousefzad@ufl.edu, rahulseng@ufl.edu, Jeremy.Dilmore@dot.state.fl.us, ranka@cise.ufl.edu}
}

\maketitle

\begin{abstract}

Urban congestion at signalized intersections leads to significant delays, economic losses, and increased emissions. Existing deep learning models often lack spatial generalizability, rely on complex architectures, and struggle with real-time deployment. To address these limitations, we propose the Temporal Graph-based Digital Twin (TGDT), a scalable framework that integrates Temporal Convolutional Networks and Attentional Graph Neural Networks for dynamic, direction-aware traffic modeling and assessment at urban corridors. TGDT estimates key Measures of Effectiveness (MOEs) for traffic flow optimization at both the intersection level (e.g., queue length, waiting time) and the corridor level (e.g., traffic volume, travel time). Its modular architecture and sequential optimization scheme enable easy extension to any number of intersections and MOEs. The model outperforms state-of-the-art baselines by accurately producing high-dimensional, concurrent multi-output estimates. It also demonstrates high robustness and accuracy across diverse traffic conditions, including extreme scenarios, while relying on only a minimal set of traffic features. Fully parallelized, TGDT can simulate over a thousand scenarios within a matter of seconds, offering a cost-effective, interpretable, and real-time solution for urban traffic management and optimization.


\end{abstract}


\section{Introduction}
The rapid urbanization and resulting traffic congestion demand smarter traffic management strategies, especially along arterial corridors where poor signal coordination exacerbates delays. In the U.S. alone, urban congestion accounts for 8.7 billion extra driving hours and \$190 billion in annual losses \cite{ttimobility2021}, while the EPA identifies idling as a major pollution source \cite{epa2022ghg}. With over 68\% of the global population projected to live in cities by 2050 \cite{unwup2018}, developing data-driven models to estimate and reduce intersection-level and corridor-level inefficiencies is increasingly urgent. Measures of Effectiveness (MOEs) such as queue length and travel time play a crucial role in adaptive traffic signal systems and smart city planning. However, most existing models rely on static, point-based predictions and fail to capture the spatiotemporal dependencies between intersections and signal phases.

While deep learning techniques like CNNs \cite{lecun1998gradient}, RNNs \cite{elman1990finding}, LSTMs \cite{hochreiter1997long}, and transformers \cite{vaswani2017attention} offer promise, they struggle with directional flows and queue spillbacks in signalized corridors. Graph Neural Networks (GNNs) \cite{li2017diffusion, wang2022hierarchical} can better represent traffic networks but face challenges with long-term dependencies and noisy data \cite{yousefzadeh2023comprehensive}. To address these issues, we propose the Temporal Graph-based Digital Twin (TGDT), which fuses Attentional GNNs with Temporal CNNs to enable efficient, non-sequential spatiotemporal learning. Our model, evaluated on a 9-intersection corridor using 50,000 hours of SUMO-simulated data \cite{876393:20293669}, is calibrated with real-world signals and behavior data. It achieves scalable, interpretable, and robust performance with minimal input requirements, offering a practical and cost-effective tool for modern traffic management.

The key contributions of this paper are summarized as follows:

\begin{itemize}
\item TGDT uses interval-based traffic volumes from collector roads to infer volumes on intervening roads, solving a graph feature imputation task with Graph Attentional Networks (GATs).

\item Traffic state matrices are assigned to nodes for intersections and edges for movement directions, forming scalable, dynamic graphs with time-evolving dependencies.

\item TGDT captures rich graph representations for modeling corridor-level MOEs (e.g., travel time) via intermediate feature fusion, while temporal CNNs are used for intersection-level MOEs (e.g., queue lengths, waiting times).

\item A sequential optimization scheme has been employed for performance improvement, but also interdependent and hierarchical learning across TGDT’s modular architecture.

\item TGDT offers a scalable, cost-effective solution for real-time urban traffic assessment with $O(1)$ sequential computation and GPU parallelization, simulating 1,000 scenarios in under one minute versus over 60 minutes with micro-simulators on 100 CPUs.
\end{itemize}

The model exhibits strong robustness and accuracy across various traffic scenarios, including extreme conditions and traffic conditions (e.g., cycle length, traffic density, and maximum green duration), using only limited traffic features. In comparison to state-of-the-art baselines, it effectively estimates multiple MOEs simultaneously. This model achieves low error rates, such as 24 seconds for travel time, 100 seconds for waiting time, and a maximum error of 4 and 1.5 vehicles for queue length and volume at every 5-minute interval estimation on a 10-mile corridor under study.

The remainder of this paper is structured as follows: Section~\ref{proposedmodels} presents the architecture of the proposed digital twin framework. Section~\ref{datagen} describes the data generation process. Section~\ref{expresults} provides a comprehensive evaluation of the model’s performance through experimental analysis. Section~\ref{related} discusses relevant prior work, and Section~\ref{conclusion} concludes the paper by summarizing key findings and outlining directions for future research.

\begin{figure*}[htbp]
        \centering
        \captionsetup{justification=raggedright,singlelinecheck=false}
        \includegraphics[scale=0.4]{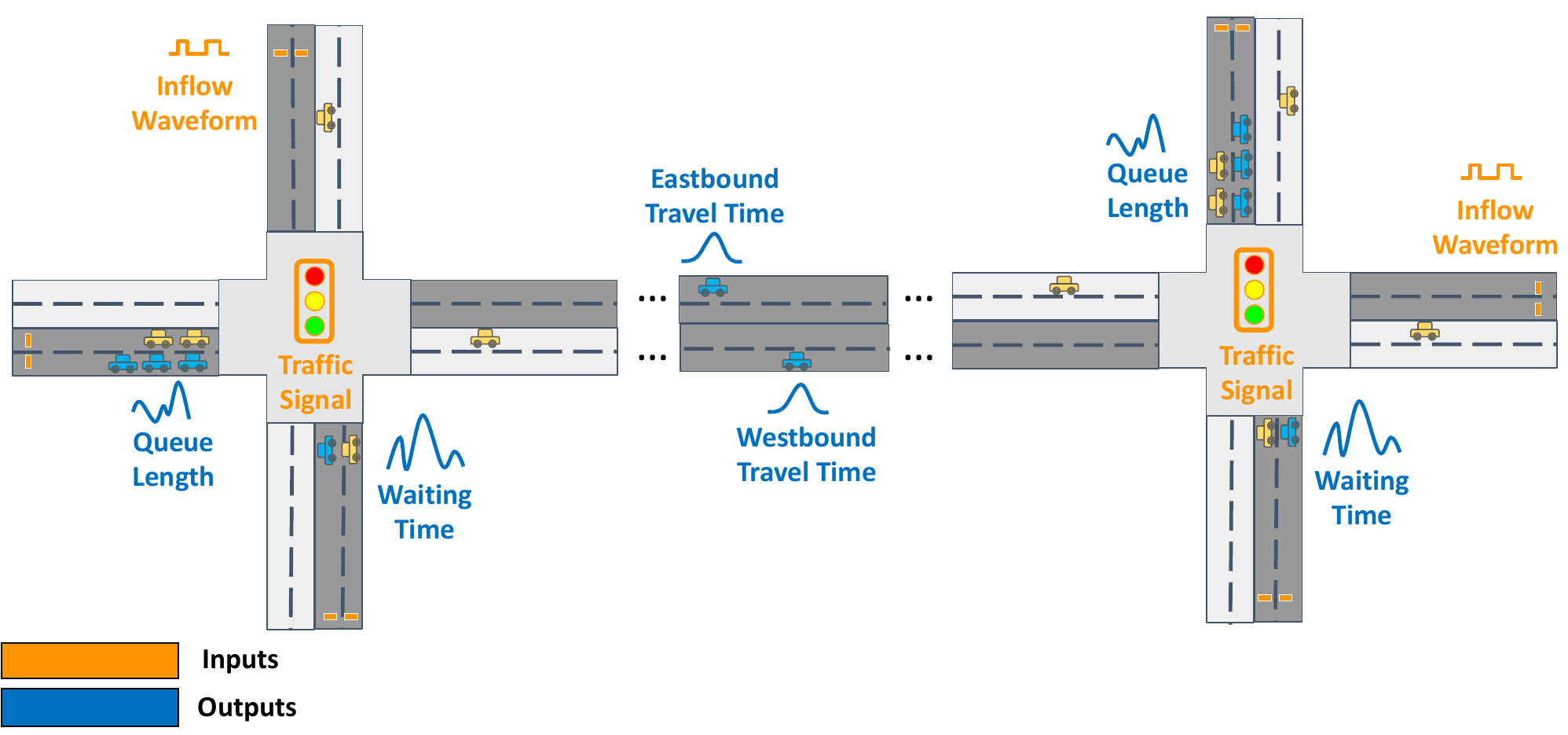}
        \renewcommand{\figurename}{Figure}
        \captionsetup{size=footnotesize}
\footnotesize
        \caption{\textbf{The inputs and outputs of the proposed urban corridor digital twin.} TGDT takes input parameters (highlighted in orange), including ingress aggregated traffic waveforms, signal timing parameters (e.g., cycle length, offset, and maximum green duration for each phase), driving behavior parameters (e.g., speed, acceleration, space cushion, lane-changing behavior), turning movement ratios, and the distances between intersections along the major corridor. It simultaneously generates multiple outputs (highlighted in blue), such as westbound travel times along the corridor, queue lengths for each lane group phase, and average waiting times for each lane group phase. The time intervals of the output time series match those of the input inflow waveforms.}
        \label{fig:corridor}
    \end{figure*}

\section{Proposed Models}
\label{proposedmodels}
In this study, we propose a structured deep digital twin framework for urban traffic corridors, capable of accurately estimating bidirectional travel time along arterial roads, as well as the maximum queue length and waiting time for both through and left-turning traffic at collector road intersections. The framework is designed to be highly flexible and generalizable, allowing it to be applied to urban corridors with arbitrary intersection counts and topologies. By leveraging the scalability of the architecture, the model can evaluate thousands of traffic scenarios within seconds, providing efficient and robust estimations of both corridor-level and intersection-level Measures of Effectiveness (MOEs).

The core of our model architecture integrates graph neural networks (GNNs) and convolutional neural networks (CNNs). The GNN component captures spatial interdependencies among intersections using a less complex yet highly effective structure, enhanced by an intermediate fusion mechanism that enables context-aware representation learning. Simultaneously, the CNN component encodes the temporal dynamics of evolving edge features, enabling the model to capture time-dependent variations in traffic patterns. This modular architecture supports a sequential optimization scheme, allowing the system to scale with any number of MOEs or corridor length without encountering gradient vanishing or exploding issues.

To ensure cost-effectiveness and practical deployment, we limit the input features to a small set of highly accessible parameters commonly collected by traffic authorities. This design choice significantly reduces the implementation burden while maintaining predictive accuracy. Moreover, the model is optimized for GPU parallelization and constant-time sequential computation, ensuring rapid inference suitable for real-time or large-scale offline applications. Overall, this digital twin framework offers a scalable, adaptable, and computationally efficient solution for advanced traffic management and corridor-level performance analysis.

An end-to-end architectural overview of the proposed Temporal Graph-based Digital Twin (TGDT) framework is illustrated in Figure~\ref{fig:model}. Following the extraction of simulation records from a microscopic traffic simulator, the raw data are processed and transformed into graph-structured representations that uniquely capture the traffic state of the corridor for each scenario (see Section~\ref{datagen} for details). The learning process along the pipeline can be formally defined as follows:

\[
\begin{split}
(1)& \quad M_{\text{inf}}: G_s(V, E, X, R_s) \rightarrow \Tilde{X} \in \mathcal{R}^{k \times p}\\
(2)& \quad M_{\text{tt}}: G_d(V, E, \Tilde{X}_c, \Tilde{X}_t, R_d) \rightarrow Y_{\text{tt}}^e, Y_{\text{tt}}^w \in \mathcal{R}^{w}\\
(3)& \quad M_{\text{ql}}: H(G_d(V, E, \Tilde{X}_c, \Tilde{X}_t, R_d)) \rightarrow Y_{\text{ql}} \in \mathcal{R}^{k \times p \times w}\\
(4)& \quad M_{\text{wt}}: H(G_d(V, E, \Tilde{X}_c, \Tilde{X}_t, R_d)) \rightarrow Y_{\text{wt}} \in \mathcal{R}^{k \times p \times w}\\
\end{split}
\]

Here, \(G_s\) and \(G_d\) denote the static and dynamic graph representations, respectively; \(V\) and \(E\) represent the sets of nodes and edges; \(X\) is the input node feature matrix; \(R_s\) and \(R_d\) correspond to time-invariant and time-varying edge feature attributes, respectively; and \(\Tilde{X}_c\) and \(\Tilde{X}_t\) represent the context and temporal embeddings derived from \(\Tilde{X}\). The operator \(H(\cdot)\) denotes the hidden spatiotemporal representation learning applied at a lower hierarchical level of the data patterns. The notations $k$, $p$, and $w$ respectively represent the number of intersections, the number of phases, and the length of the time series in input/output of the modules.

The four modules of TGDT represent: (1) Inflow module $M_{inf}$: intervening traffic volume imputation, (2) Travel time module $M_{tt}$: Corridor-level travel time estimation, (3) Queue length module $M_{ql}$: Intersection-level and phase-wise maximum queue length estimation, and (4) Waiting time module $M_{wt}$: Intersection-level and phase-wise waiting time estimation.\footnote{Source code: https://github.com/NSH2022/Temporal-Graph-Based-Digital-Twin}

\begin{figure*}[htbp]
        \centering
        \captionsetup{justification=raggedright,singlelinecheck=false}
        \includegraphics[scale=0.4]{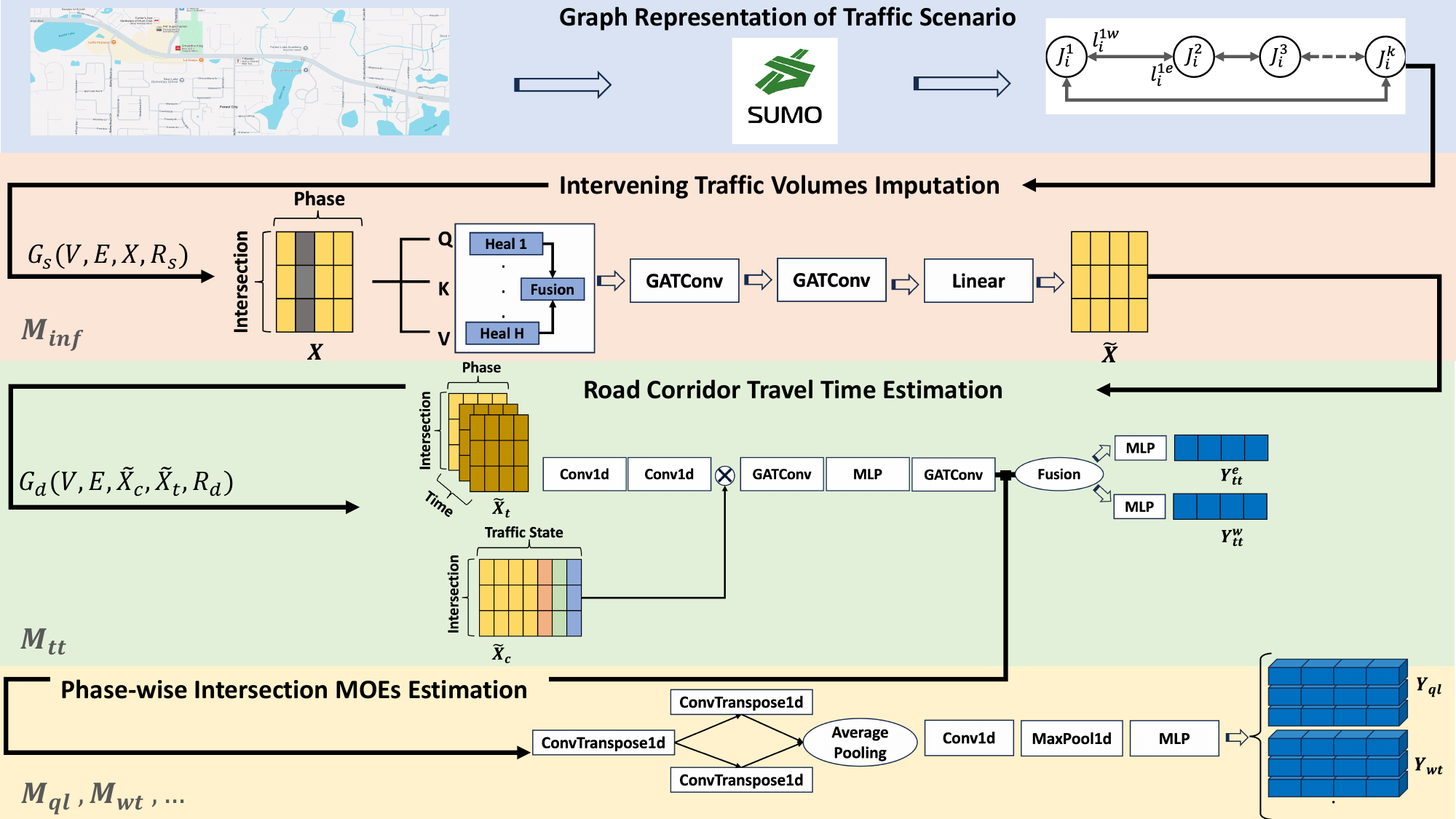}
        \renewcommand{\figurename}{Figure}
        \captionsetup{size=footnotesize}
\footnotesize
        \caption{\textbf{Overview of TGDT framework.} This diagram illustrates the architecture of our proposed Digital Twin for urban corridors, which consists of three main modules. Simulation records, extracted from the logs of a microscopic traffic simulator, are transformed into graph-structured data that uniquely represent the corridor’s traffic state for each scenario. The inflow module (\(M_{\text{inf}}\)) performs a graph imputation task to reconstruct 2D traffic volumes on every intermediate road segment. The travel time module (\(M_{\text{tt}}\)) carries out a graph regression task to estimate bidirectional corridor-level travel time series. Finally, the queue length (\(M_{\text{ql}}\)) and waiting time (\(M_{\text{wt}}\)) modules apply temporal convolution and deconvolution operations on the spatiotemporal representations learned by \(M_{\text{tt}}\), producing 3D outputs for maximum queue length and waiting time. These estimates are generated at the intersection and phase levels for each lane group associated with a specific movement phase.
}
        \label{fig:model}
    \end{figure*}

The inflow module ($M_{inf}$) extends our previous work \cite{10919173} from intersection-level to corridor-scale modeling. It consists of a self-attention module \cite{vaswani2017attention} followed by two Graph Attention Network (GAT) layers with 4 and 1 attention heads, respectively, followed by a fully connected layer, all activated by ReLU. This module imputes traffic volumes at the location of the inflow loop detector simulated at 500 meters upstream of each intersection. The input traffic simulation graph is constructed as a directed acyclic graph with 8 nodes (intersections) and 16 edges (directional road segments). Nodes are attributed with 5-minute aggregated inflow volumes, while edges carry 14 features, including inter-intersection distance, turning movement counts, and traffic density.

The travel time module ($M_{tt}$) follows a similar architecture with two GAT layers (4 and 1 attention heads), embedding 14-dimensional node features into a 64-dimensional hidden representation. It incorporates an edge MLP submodule and two fully connected layers to generate travel time predictions in both directions of the corridor. All layers use the ReLU activation function. The second GAT layer processes updated edge features from the MLP submodule, alongside node embeddings from the first GAT layer, while maintaining the same edge connectivity. The edge MLP submodule models the time-evolving behavior of linkages between intersections, representing the system as a dynamic graph, enabling the use of richer node features and temporally-aware edge features. An intermediate fusion technique merges node and edge embeddings into a unified graph representation. This design enables the model to effectively address the graph regression task of estimating travel times for both directions of arterial roads at 5-minute intervals.

The queue length module (\(M_{\text{ql}}\)) and waiting time module (\(M_{\text{wt}}\)) share the same architecture, which is based on the structure of \(M_{\text{tt}}\). However, since they aim to estimate high-dimensional outputs at multiple points along the corridor, the representation learned by \(M_{\text{tt}}\) is first abstracted before the fusion of graph features. This abstracted representation is then passed through additional layers that combine convolutional and transposed convolutional operations for multi-channel temporal feature learning from 1D multivariate inputs. Specifically, two ConvTranspose1d layers with kernel size of $w$ and stride of 1 upscale the temporal features for each direction of movement. The upsampled outputs are then passed through a CNN encoder composed of successive Conv1d, MaxPool1d, and fully connected layers, concluding with a ReLU activation to extract high-level representations.

\subsection{Sequential Optimization Technique}

We implement our framework using the PyTorch machine learning framework and the PyTorch Geometric library. A sequential optimization strategy is adopted during training, where the parameters of individual modules are updated one at a time rather than optimizing the entire network jointly. This method is particularly effective when modules are interdependent or serve distinct roles, allowing each to be fine-tuned independently for better convergence and overall performance.

To support this, we define three independent Adam optimizers with distinct learning rates and mean squared error (MSE) loss function to guide updates for the inflow, travel time, queue length, and waiting time modules in a prioritized order. The overall loss is composed of four task-specific components, each optimized independently. Notably, the travel time loss is computed as the sum of directional losses (eastbound and westbound) across the corridor.

This modular and sequential training framework ensures effective learning for each component, leading to more accurate modeling of traffic dynamics and enabling scalability to additional Measures of Effectiveness (MOEs). The architecture also enhances interpretability and robustness, as the estimation of lower-hierarchy MOEs logically depends on the outputs of the model produced for higher-hierarchy MOEs, underscoring the importance of the sequential optimization approach employed in this structured framework.

\section{Traffic Data Generation}
\label{datagen}

This section outlines the dataset generation and preprocessing steps, divided into three main phases: Traffic Simulation, Log Extraction, and Graph Data Construction.

\subsection{Traffic Simulation}
Traffic data was generated using the SUMO (Simulation of Urban MObility) \cite{876393:20293669} micro-simulator, modeling eight east-west intersections along Florida's SR 436 arterial road with realistic configurations and parameters. Over 50,000 hours of simulation were executed in parallel using Python's ThreadPoolExecutor. Origin-Destination (OD) matrices were estimated using real-world ATSPM loop detectors and WEJO GPS data, then used to generate routes via SUMO’s od2trips tool. The resulting dataset, Real-TMC, is based on both inferred and randomized OD matrices.

\subsection{Log Extraction}
Each simulation run produces XML log files containing vehicle trajectories and metrics. Travel times across the arterial corridor are computed from SUMO’s Floating Car Data (FCD), using vehicle entry and exit times across intersections, specifically for East-West and West-East directions. Only vehicles completing the full route are included in the final travel time metrics.

\subsection{Graph Data Construction}
\label{graphrep}

The dataset contains 50,000 graph samples, each modeled as a bidirectional, acyclic graph $G(V, E)$ with $|V| = 8$ and $|E| = 16$, representing 8 intersections and 16 directed road segments.
The acyclic design prevents inefficiencies like infinite loops during traversal and learning.
Node features are direction-agnostic; edge features $R \in \mathcal{R}^{16 \times 19}$ are direction-specific, encoding roadway traits like distance, turning counts, traffic density, and driving behavior.
Time-varying features on edges support sequence modeling and reflect traffic fluctuations over time.
Two graph types are used: static and dynamic, to adapt to spatial and temporal learning needs.
The static graph is $G_s(V, E, X, R_s)$, where $X \in \mathcal{R}^{8 \times 8}$ represents inflow volumes from loop detectors, masked to exclude arterial traffic.
A module estimates missing values and produces $\Tilde{X}$, reused in the dynamic graph $G_d(V, E, \Tilde{X}_c, \Tilde{X}_t, R_d)$.
The dynamic graph includes a node tensor $X_c \in \mathcal{R}^{8 \times 14 \times 10}$ capturing 14 time-series features across 10 time steps.
These features include signal timing plans (e.g., cycle length, offset, and max green time).
Target variables are aligned as time series with uniform intervals, ensuring consistent training and supervision across tasks.

\section{Experimental Results}
\label{expresults}
In this section, we evaluate the performance of our proposed  Temporal Graph-based Digital Twin (TGDT) in estimating traffic volumes, queue lengths, waiting times, and travel times within urban traffic corridors. To assess the sensitivity of model outputs to the temporal resolution of input features, we compare the original model, \textbf{TGDT}, trained on 5-minute aggregated data, with a variant of it, \textbf{TGDT-short}, trained on input data aggregated into 1-minute intervals. Additionally, we benchmark both models against previously published approaches that achieve state-of-the-art accuracy and interpretability.

TGDT simultaneously models three primary Measures of Effectiveness (MOEs) for every intersection and for each phase, direction, and bound of movement. We evaluate the accuracy of intersection-wise and phase-wise queue length estimation and inflow waveform imputation by our model by comparing them to the results obtained by \textbf{MTDT} digital twin \cite{10920162} and \textbf{GAT-AE} graph autoencoder \cite{10919173} trained on input data with 5-second resolution. For the estimation of bidirectional travel time, we assess our model against \textbf{FDGNN} in \cite{yousefzadeh2024dynamic}, which can approximate normal travel time distributions from travel time histograms.

The dataset is partitioned into 70\% for training, with the remaining 30\% split equally: 15\% for hyperparameter tuning and 15\% for model evaluation. To ensure a comprehensive and rigorous evaluation of our proposed architecture, model performance to estimate each Measure of Effectiveness (MOE) is compared against those obtained from others at the indicated size of aggregation windows $W$ of time series using several widely accepted metrics:

\begin{itemize}

    \item \textbf{Normalized Root Mean Squared Error (NRMSE)}:
    \[
    \text{NRMSE} = \frac{\sqrt{\frac{1}{n} \sum_{i=1}^{n} \left( y^{\text{true}}_i - y^{\text{pred}}_i \right)^2}}{\max(y_{\text{true}}) - \min(y_{\text{true}})}
    \]

    \item \textbf{Hellinger Distance (HLD)}:
    \[
    \text{HLD} =  \frac{1}{\sqrt{2}} \sqrt{ \sum_{i=1}^{n} \left( \sqrt{y^{\text{true}}_i} - \sqrt{y^{\text{pred}}_i} \right)^2}
    \]

    \item \textbf{Normalized Earth Mover’s Distance (EMD)}:
    \small
    \[
    \text{EMD} = \frac{\sum_{i=1}^{n} \sum_{j=1}^{n} f(y^{\text{true}}_i \rightarrow y^{\text{pred}}_j) \cdot (|y^{\text{true}}_i - y^{\text{pred}}_j|)}{\sum_{i=1}^{n} \sum_{j=1}^{n} f(y^{\text{true}}_i \rightarrow y^{\text{pred}}_j)}
    \]
    where \( \sigma_{\text{true}} \) and \( \sigma_{\text{pred}} \) are the standard deviations of the true and predicted distributions, respectively.
    
    \item \textbf{Mean Absolute Percentage Error (MAPE)}:
    \[
    \text{MAPE} = \frac{1}{n} \sum_{i=1}^{n} \left| \frac{y^{\text{true}}_i - y^{\text{pred}}_i}{y^{\text{true}}_i} \right| \times 100
    \]
    
    where $i$ represents the time step in the actual and predicted time series, and $f_{ij}$ denotes the optimal flow from the actual value at time step $i$ to the predicted value at time step $j$, determined by solving a transportation optimization problem.
\end{itemize}

Figure~\ref{fig:results} presents a visual comparison between the actual (red) and predicted (green) time series of three primary MOEs (i.e, travel time, queue length, and waiting time) simultaneously generated by our model for a sample intersection at a randomly selected movement phase, direction, and bound.

\begin{figure*}[htbp]
        \centering
        \captionsetup{justification=raggedright,singlelinecheck=false}
        \includegraphics[scale=0.4]{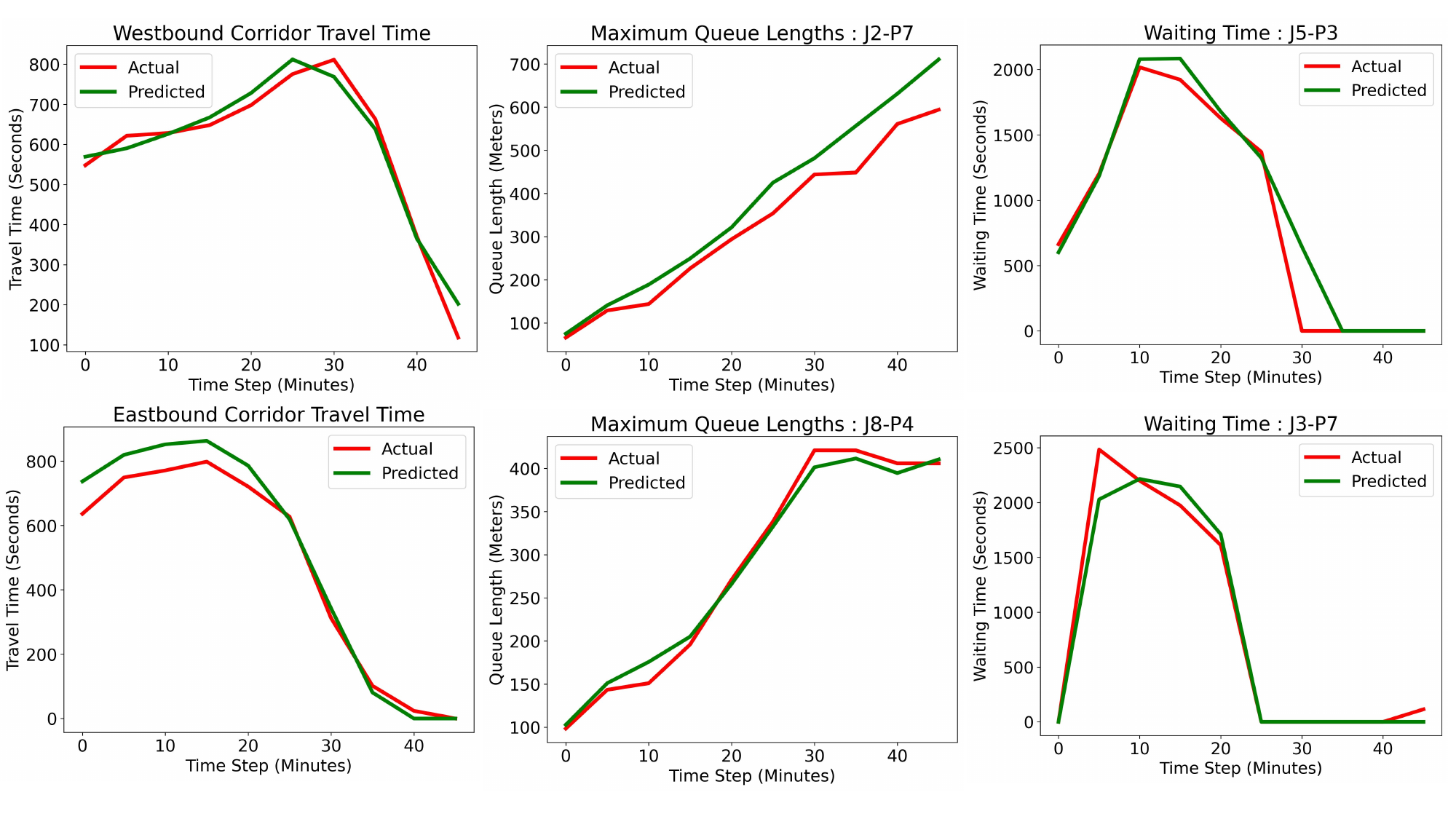}
        \renewcommand{\figurename}{Figure}
        \captionsetup{size=footnotesize}
\footnotesize
        \caption{\textbf{Visualization outputs of TGDT for a randomly selected traffic scenario.} Comparison of actual (red) and predicted (green) curves for several Measures of Effectiveness (MOEs). TGDT takes ingress (inflow) traffic volumes and various parameters, such as traffic signal plans and driving behaviors, to accurately simulate bidirectional travel times throughout urban traffic corridors. It also estimates multi-directional maximum queue lengths and waiting times within a given time interval. The model first imputes intermediate inflow traffic volumes between intersections, then analyzes traffic performance across all intersections ($J_1, ..., J_8$) and all phase lane groups ($P_1, ..., P_8$) simultaneously. }
        \label{fig:results}
    \end{figure*}

Figure~\ref{fig:kde} demonstrates the difference between estimated and actual distribution of corridor travel times at a certain time step for Eastbound and Westbound traffic directions. Kernel Density Estimates (KDEs) show that the predicted distributions closely match the actual ones, with minimal deviations in the tails for the Eastbound direction (with sparse traffic) and in the peak for the Westbound (with intense traffic) direction. Here, as shown in the 2D kernel density plots of mean ($\mu$) versus standard deviation ($\sigma$), the clustering and spread of the estimated travel time statistics closely match those of the actual distribution. This alignment reflects strong predictive accuracy and low distributional divergence. The visual similarity accounts for the low Normalized Earth Mover’s Distance (EMD = 0.04), indicating minimal spatial adjustment needed to align the distributions. However, minor variations in distributional shape across directions contribute to different Hellinger Distances (HLD = 0.08 vs. HLD = 0.3), consistent with the error values reported in Table~\ref{table:error}.

\begin{figure*}[htbp]
        \centering
        \captionsetup{justification=raggedright,singlelinecheck=false}
        \includegraphics[scale=0.37]{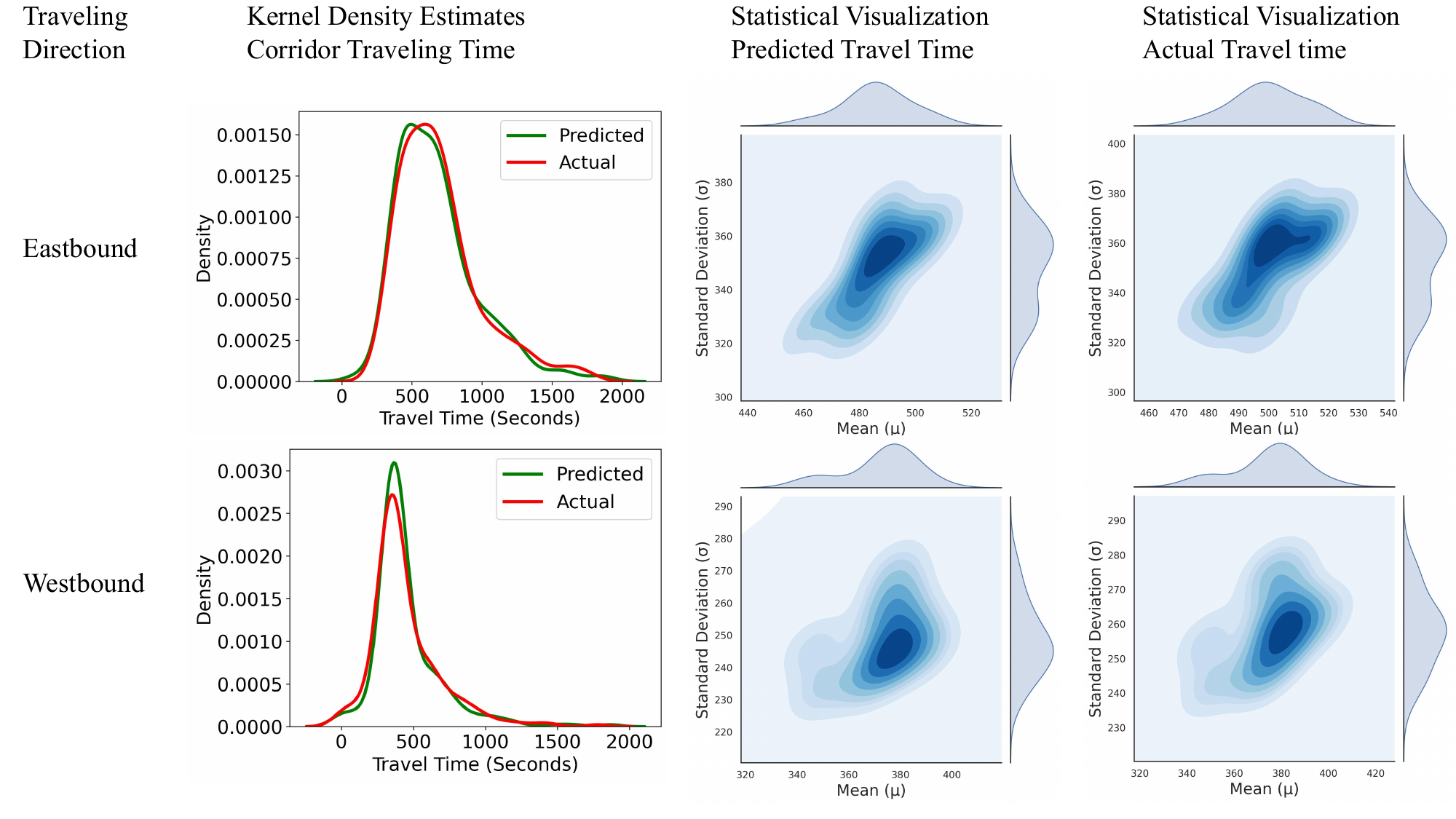}
        \renewcommand{\figurename}{Figure}
        \captionsetup{size=footnotesize}
\footnotesize
        \caption{\textbf{Predicted versus actual travel time distributions.} This figure shows the Kernel Density Estimates (left column) and the joint distribution of travel time characteristics (right column) across the target corridor at a certain time step $T=15$ min. The results suggest that the predicted behavior closely matches the actual travel time in terms of both the average ($\mu$, slightly underestimated) and variability ($\sigma$, slightly overestimated).}
        \label{fig:kde}
    \end{figure*}
Further, to comprehensively assess the robustness of the model’s performance across varying traffic dynamics and signal control conditions, we partition the test set into subsets using predefined thresholds across three key dimensions:

\subsection{Effect of Cycle Length}
To examine the sensitivity of model performance to signal timing, we group test samples based on signal cycle length:
\begin{itemize}
    \item \textbf{Low:} Cycle length $<$ 160 seconds
    \item \textbf{Medium:} 160 $\leq$ Cycle length $<$ 200 seconds
    \item \textbf{High:} Cycle length $\geq$ 200 seconds
\end{itemize}
This allows us to evaluate how the model performs under different signal timing strategies.

\subsection{Effect of Traffic Volume}
To analyze the impact of varying demand levels, we categorize samples based on the number of vehicles completing a corridor trip in a specific direction:
\begin{itemize}
    \item \textbf{Low:} Traffic volume $<$ 700 vehicles
    \item \textbf{Medium:} 700 $\leq$ Traffic Volume $<$ 900 vehicles
    \item \textbf{High:} Traffic volume $\geq$ 900 vehicles
\end{itemize}
This highlights model robustness under low, medium, and high traffic demand conditions.

\subsection{Effect of Maximum Green Duration Percentage}
To assess the model’s responsiveness to traffic control policies, we divide samples based on the ratio of maximum green time allocated to the through phase of the corridor relative to the cycle length:
\begin{itemize}
    \item \textbf{Low:} Maximum green time $<$ 25\% of Cycle length
    \item \textbf{Medium:} 25\% $\leq$ Maximum green time $<$ 50\%
    \item \textbf{High:} Maximum green time $\geq$ 50\%
\end{itemize}
This allows us to examine how green time allocation influences the model’s predictive performance.

As shown by the error values in Table~\ref{table:error}, the proposed TGDT model demonstrates superior performance in estimating both travel time and waiting time while trained on time series aggregated on time intervals of $w = 5$ minutes. A point-wise comparison between predicted and actual values is conducted using multiple evaluation metrics. MAPE shows low errors—approximately 30 seconds for travel time and 100 seconds for waiting time—indicating precise point-wise predictions. While distribution-based metrics, including a Hellinger Distance (HLD) of 0.08 and a Normalized Earth Mover’s Distance (EMD) of 0.04, suggest that the model effectively captures the overall shape and alignment of the actual time series. Although the waiting time series shows slightly higher variation, both distributions largely align, reflecting the model’s strength in learning temporal patterns.

Beyond travel and waiting time estimation, TGDT also performs well in multi-directional queue length predictions and is adaptable to various time resolutions. Both TGDT and its short-interval variant outperform the MTDT model, even when the latter uses finer-grained input. The TGDT model maintains robustness under diverse traffic dynamics, with cycle length and traffic volume exerting minimal effects on overall performance. Moreover, a higher share of maximum green time improves most performance metrics, highlighting the importance of adaptive signal timing and interval selection to ensure reliable and scalable traffic management solutions.

\begin{table*}[!htbp]
    \centering
    \resizebox{0.99\textwidth}{!}
    {
    \begin{tabular}{@{}lccccccccccccc@{}}
        \toprule
        \multirow{2}{*}{} & \multicolumn{1}{c}{} & \multicolumn{4}{c}{\textbf{TGDT ($w = 5 \: min$)}} & \multicolumn{4}{c}{\textbf{TGDT-Short ($w = 1 \: min$)}} & \multicolumn{4}{c}{\textbf{FDGNN ($w = 5 \: min$)}} \\ 
        \cmidrule(lr){3-6} \cmidrule(lr){7-10} \cmidrule(lr){11-14}
        Experiment & Level & MAPE & EMD& HLD & NRMSE & MAPE & EMD& HLD & NRMSE & MAPE & EMD& HLD & NRMSE \\ \midrule
        \multicolumn{14}{c}{\textbf{Travel Time}} \\ \midrule
        \multirow{3}{*}{Cycle Length} 
        & Low    & 0.0327 & 0.0842 & 0.0913 & 0.0431 & 0.0263 & 0.1434 & 0.1263 & 0.0767 &  0.0276& 0.0526 &  0.1032&  0.0685\\ 
        & Medium & 0.0270 & 0.0444 & 0.0717 & 0.0325 & 0.0338 & 0.0968 & 0.1081 & 0.0484 &  0.0255& 0.0671 &  0.0957&  0.0658\\ 
        & High   & 0.0257 & 0.0373 & 0.0723 & 0.0296 & 0.0320 & 0.0756 & 0.1008 & 0.0381 &  0.0230&  0.1032&  0.0868&  0.0548\\ 
        \midrule
        \multirow{3}{*}{Traffic Volume} 
        & Low    & 0.0160 & 0.0483 & 0.08453 & 0.0228 & 0.0323 & 0.1026 & 0.1198 & 0.0397 &  0.0272&  0.0729&  0.1019&  0.0627\\ 
        & Medium & 0.0447 & 0.0582 & 0.0729 & 0.0296 & 0.0266 & 0.0824 & 0.1035 & 0.0366 &  0.0255& 0.0543 &  0.0957&  0.0573\\ 
        & High   & 0.0496 & 0.0676 & 0.0712 & 0.0291 & 0.0405 & 0.1218 & 0.1122 & 0.0490 &  0.0217& 0.0481 &  0.0817&  0.0536\\ 
        \midrule
        \multirow{3}{*}{Maximum Duration \%} 
        & Low    & 0.0509 & 0.0594 & 0.0708 & 0.0344 & 0.0148 & 0.0521 & 0.0931 & 0.0415 &  0.0208& 0.1298 &  0.0788&  0.0734\\ 
        & Medium & 0.0279 & 0.0468 & 0.07973 & 0.0254 & 0.0314 & 0.0974 & 0.1137 & 0.0369 &  0.0257& 0.0555 &  0.0962&  0.0572\\ 
        & High   & 0.0381 & 0.1389 & 0.0973 & 0.0665 & 0.0677 & 0.1376 & 0.1081 & 0.0611 &  0.0195& 0.1489 &  0.0747&  0.0661\\ 
        \midrule
        & Total  & \textbf{0.0248} & \textbf{0.0413} & \textbf{0.0877} & \textbf{0.0347} & 0.0305 & 0.0932 & 0.1186 & 0.0460 &  0.0258&0.0558  &  0.0967&  0.0563\\ 
        \midrule
        \toprule
        \multirow{2}{*}{.} & \multicolumn{1}{c}{} & \multicolumn{4}{c}{\textbf{TGDT ($w = 5 \: min$)}} & \multicolumn{4}{c}{\textbf{TGDT-Short ($w = 1 \: min$)}} & \multicolumn{4}{c}{\textbf{MTDT ($w = 5 \: sec$)}} \\ 
        \cmidrule(lr){3-6} \cmidrule(lr){7-10} \cmidrule(lr){11-14}
        Experiment & Level & MAE& MSE& RMSE & NRMSE & MAE& MSE& RMSE & NRMSE & MAE& MSE& RMSE & NRMSE \\ \midrule
        \multicolumn{14}{c}{\textbf{Maximum Queue Length}} \\ \midrule
        \multirow{3}{*}{Cycle Length} 
        & Low    & 20.099 & 1729.3 & 41.585 & 0.0419 & 17.972 & 1411.6 & 37.571 & 0.0379 & 29.993 &23974  & 154.83 & 0.1039 \\ 
        & Medium & 21.039 & 1761.3 & 41.968 & 0.0423 & 18.952 & 1545.5 & 39.313 & 0.0396 & 38.354 & 25377 &  159.30& 0.0808 \\ 
        & High   & 22.016 & 1859.5 & 43.122 & 0.0435 & 20.47 & 1739.7 & 1739.7 & 0.0421 & 40.436 &  22171&148.90  & 0.0752 \\ 
        \midrule
        \multirow{3}{*}{Traffic Volume} 
        & Low    & 18.438 & 1422.0 & 37.709 & 0.0380 & 15.842 & 1135.9 & 33.703 & 0.0340 &  24.253&12934  & 113.73 & 0.0575 \\ 
        & Medium & 24.112 & 2211.9 & 47.031 & 0.0474 & 22.832 & 2064.5 & 45.437 & 0.0458 & 51.996 & 36687 & 191.54 & 0.0999 \\ 
        & High   & 25.797 & 2532.2 & 50.321 & 0.0508 & 24.303 & 2182.6 & 46.718 & 0.0471 &  49.383& 50270 &  224.21& 0.1431 \\ 
        \midrule
        \multirow{3}{*}{Maximum Duration \%} 
        & Low    & 21.332 & 1722.4 & 41.502 & 0.0419 & 18.665 & 1453.4 & 38.123 & 0.0384 & 29.993 & 23974 & 154.83 & 0.1039 \\ 
        & Medium & 20.983 & 1789.0 & 42.297 & 0.0427 & 19.031 & 1555.6 & 39.441 & 0.0398 & 38.354 & 25377 & 159.30 & 0.0808 \\ 
        & High   & 24.220 & 2643.6 & 51.416 & 0.0519 & 20.077 & 1701.5 & 41.249 & 0.0416 & 40.436 & 22171 &  148.90& 0.0752 \\ 
        \midrule
        & Total  & 21.152 & 1798.9 & 42.414 & 0.0428 & \textbf{18.921} & \textbf{1540.3} & \textbf{39.246} & \textbf{0.0396} &35.8580  & 22440 &149.80  & 0.0760\\ 
        \midrule
        \toprule
        \multirow{2}{*}{.} & \multicolumn{1}{c}{} & \multicolumn{4}{c}{\textbf{TGDT ($w = 5 \: min$)}} & \multicolumn{4}{c}{\textbf{TGDT-Short ($w = 1 \: min$)}} & \multicolumn{4}{c}{\textbf{GAT-AE ($w = 5 \: sec$)}} \\ 
        \cmidrule(lr){3-6} \cmidrule(lr){7-10} \cmidrule(lr){11-14}
        Experiment & Level & MAE& MSE& RMSE & NRMSE & MAE& MSE& RMSE & NRMSE & MAE& MSE& RMSE & NRMSE \\ \midrule
        
        \multicolumn{14}{c}{\textbf{Intervening Traffic Volume}} \\ \midrule
        \multirow{3}{*}{Cycle Length} 
        & Low    & 5.2414 & 105.84 & 10.288 & 0.0088 & 17.834 & 1757.5 & 41.923 & 0.0372 & 1.1785 & 8.3624 & 2.8917 & 0.0788 \\ 
        & Medium & 5.2789 & 107.35 & 10.361 & 0.0089 & 16.986 & 1609.7 & 40.121 & 0.0337 & 1.1693 & 8.7187 &  2.9527& 0.0798 \\ 
        & High   & 5.3168 & 109.49 & 10.464 & 0.0091 & 16.841 & 1558.9 & 39.483 & 0.0348 & 1.1477 & 8.6280 & 2.9373 & 0.0793 \\ 
        \midrule
        \multirow{3}{*}{Traffic Volume} 
        & Low    & 4.6576 & 83.721 & 9.1499 & 0.0078 & 15.524 & 1184.0 & 34.409 & 0.0303 & 1.0911 & 7.2249 & 2.6879 & 0.0726 \\ 
        & Medium & 6.0597 & 137.71 & 11.735 & 0.0101 & 19.092 & 2157.4 & 46.447 & 0.0391 & 1.2672 &  10.287& 3.2074 & 0.0866 \\ 
        & High   & 6.3231 & 145.45 & 12.060 & 0.0115 & 21.794 & 2738.9 & 52.335 & 0.0456 & 1.2842 & 10.875 & 3.2978 &  0.0891\\ 
        \midrule
        \multirow{3}{*}{Maximum Duration \%} 
        & Low    & 5.4501 & 114.59 & 10.705 & 0.0105 & 15.884 & 1293.7 & 35.968 & 0.0378 & 1.1400 & 8.3422 & 2.8883 & 0.0825 \\ 
        & Medium & 5.2702 & 107.16 & 10.352 & 0.0088 & 17.324 & 1664.8 & 40.802 & 0.0343 & 1.1674 &  8.5644& 2.9265  & 0.0790 \\ 
        & High   & 5.2633 & 107.31 & 10.359 & 0.0113 & 17.764 & 1760.5 & 41.958 & 0.0396 & 1.1977 & 9.3350 &  3.0553& 0.0848 \\ 
        \midrule
        & Total  & \textbf{5.2843} & \textbf{107.24} & \textbf{10.356} & \textbf{0.0086} & 17.196 & 1626.7 & 40.332 & 0.0351 & 1.1815 & 7.9688 & 2.6230 &0.0622  \\ 
        \midrule
        \toprule
        \multirow{2}{*}{.} & \multicolumn{1}{c}{} & \multicolumn{4}{c}
        {\textbf{TGDT ($w = 5 \: min$)}} & \multicolumn{4}{c}{\textbf{TGDT-Short ($w = 1 \: min$)}} & \multicolumn{4}{c}{-} \\ 
        \cmidrule(lr){3-6} \cmidrule(lr){7-10} \cmidrule(lr){11-14}
        Experiment & Level & MAPE & EMD& HLD & NRMSE & MAPE & EMD& HLD & NRMSE & MAPE & EMD& HLD & NRMSE \\ \midrule
        
        \multicolumn{14}{c}{\textbf{Waiting Time}} \\ \midrule
        \multirow{3}{*}{Cycle Length} 
        & Low    & 0.0495 & 0.0487 & 0.2946 & 0.0239 & 0.1599 & 0.0430 & 0.3789 & 0.0171 &  &  &  &  \\ 
        & Medium & 0.0464 & 0.0387 & 0.2815 & 0.0264 & 0.1527 & 0.0417 & 0.3712 & 0.0202 &  &  &  &  \\ 
        & High   & 0.0669 & 0.0422 & 0.2799 & 0.0290 & 0.1425 & 0.0404 & 0.3799 & 0.0213 &  &  &  &  \\ 
        \midrule
        \multirow{3}{*}{Traffic Volume} 
        & Low    & 0.0573 & 0.0432 & 0.2867 & 0.0228 & 0.1708 & 0.0441 & 0.3862 & 0.0174 &  &  &  &  \\ 
        & Medium & 0.0499 & 0.0374 & 0.2831 & 0.0305 & 0.1320 & 0.0369 & 0.3622 & 0.0243 &  &  &  &  \\ 
        & High   & 0.0604 & 0.0451 & 0.2977 & 0.0359 & 0.1442 & 0.0447 & 0.3782 & 0.0295 &  &  &  &  \\ 
        \midrule
        \multirow{3}{*}{Maximum Duration \%} 
        & Low    & 0.0775 & 0.0431 & 0.2708 & 0.0328 & 0.1427 & 0.0417 & 0.3781 & 0.0273 &  &  &  &  \\ 
        & Medium & 0.0541 & 0.0403 & 0.2867 & 0.0222 & 0.1516 & 0.0407 & 0.3766 & 0.0163 &  &  &  &  \\ 
        & High   & 0.0712 & 0.0388 & 0.3099 & 0.0508 & 0.1752 & 0.0497 & 0.3787 & 0.0354 &  &  &  &  \\ 
        \midrule
        & Total  & \textbf{0.0551} & \textbf{0.0403} & \textbf{0.2889} & \textbf{0.0438} & 0.1501 & 0.0404 & 0.3747 & 0.0344 &  &  &  &  \\ 
        
        \bottomrule
    \end{tabular}%
    }
    \renewcommand{\tablename}{Table}
    \caption{\textbf{Performance evaluation of TGDT.} The performance of the proposed urban corridor digital twin is evaluated using both regular and short-length time interval windows, across multiple scenario-specific groupings based on cycle length, traffic volume density, and the ratio of maximum green duration. The results for each Measure of Effectiveness (MOE) are compared against those obtained from state-of-the-art models at the indicated size of aggregation windows. Model outputs are assessed using appropriate evaluation metrics, including Mean Absolute Percentage Error (MAPE), Normalized Earth Mover’s Distance (EMD), Hellinger distance (HLD), normalized root mean square error (NRMSE), root mean square error (RMSE), and mean absolute error (MAE).}
    \label{table:error}
\end{table*}

\section{Related Work}
\label{related} 
Advancements in Artificial Intelligence (AI), particularly in deep learning, have paved the way for data-driven, scalable, and adaptive solutions in traffic management. Traffic performance assessment by Estimating Measures of Effectiveness (MOEs) such as travel time, delay, and queue length in urban traffic corridors and networks has attracted increasing attention with the advances in deep learning.

In this realm of study, Long Short-Term Memory (LSTM) networks \cite{ma2015long} and their variants, including Bidirectional LSTM (BiLSTM) \cite{siami2019performance} and Convolutional LSTM (ConvLSTM) \cite{lin2020self}, effectively capture spatiotemporal patterns for corridor-level traffic measures. However, they lack native support for spatial dependencies and suffer from limited parallelism and vanishing gradients on long corridors. 

Graph Neural Networks (GNNs) \cite{yu2017spatio} address spatial correlations in non-Euclidean traffic networks. Graph convolutions are integrated with gated temporal convolutions in \cite{yu2017spatio} for modeling the temporal dynamics of traffic flow in road networks, however, it falls short in adapting to dynamic graph structure in real-world urban traffic systems. Models like ST-GCN \cite{wang2022hierarchical} and DCRNN \cite{li2017diffusion} estimate travel metrics using graph-based representations, but struggle with long-term temporal modeling, input noise sensitivity, and computational demands. Graph WaveNet~\cite{wu2019graph} excels at learning adaptive adjacency matrices and long-range dependencies through dilated convolutions. STGODE~\cite{fang2021spatial} models continuous-time dynamics using neural ordinary differential equations (ODEs) on dynamic graphs, enabling fine-grained temporal representation. TGNN~\cite{rossi2020temporal} captures event-driven temporal transitions on dynamic graphs and is well-suited for asynchronous data. However, none of these models supports modular decomposition aligned with real-world traffic subsystems. In contrast, our TGDT employs a modular architecture inspired by digital twin principles, enabling structured multi-task outputs and allowing it to model multiple interrelated variables while maintaining structural transparency and interpretability across spatio-temporal and causal dimensions.

CNN-RNN hybrids \cite{zhang2017deep} extract spatial and temporal features, i.e., \cite{li2017graph} uses residual networks to predict citywide crowd flows by modeling complex spatial and temporal dependencies. However, they still suffer from noise sensitivity, rigid grid assumptions, and complex architectures requiring extensive tuning. Transformer-based models \cite{vaswani2017attention} improve temporal dynamics via attention, though they lack topological spatial awareness and demand large-scale data.

To overcome these limitations, we propose a Temporal Graph-based Digital Twin (TGDT) that combines Temporal CNNs with Attentional GNNs (GATs). This integration supports efficient spatiotemporal encoding, robust performance in noisy or unseen scenarios, and simplified, parallelizable training without sacrificing spatial structural modeling.

\section{Conclusions and Future Work}
\label{conclusion}

This paper tackles the challenge of designing a deep learning framework that serves as a digital twin for urban traffic corridors, offering a precise, robust, reliable, and scalable solution for evaluating traffic performance across various metrics and spatiotemporal scales. Our model integrates convolutional operations on multivariate time series for localized temporal encoding, applied on top of spatiotemporal representations learned through Graph Neural Networks. This enables the reconstruction of a Temporal Graph Attention Network that operates on dynamic traffic graphs with evolving edge features.

We introduce a fusion-based approach and a sequential optimization scheme that allows the model to scale effectively to any number of Measures of Effectiveness (MOEs) without compromising the accuracy of individual tasks or encountering gradient interference. The design remains cost-efficient, requiring minimal input features, and computationally optimized for parallel execution.

The model demonstrates strong resilience and robustness, as validated through a series of controlled experiments assessing its sensitivity to traffic signal cycle length, traffic volume density, and green time allocation. These experiments show the model's effectiveness in accurately estimating bidirectional arterial travel time distributions, even under counterfactual and randomly generated traffic scenarios, using only a limited set of accessible traffic features. For the target arterial road (Florida’s SR 436, which is 8 miles long with approximately 20-minute westbound and 15-minute eastbound travel times and includes 8 signalized intersections), the model achieves low error rates: 24 seconds for arterial travel time estimation and 100 seconds for average waiting time. It also estimates queue lengths and directional traffic volumes with a maximum error of 4 and 1.5 vehicles, respectively, at any 5-minute interval.

We believe the introduced approach is promising in real-time traffic signal optimization and adaptive control, contributing to the advancement of smart city infrastructure through efficient and scalable traffic management at the corridor level. A promising application of this work could efficiently find optimal signal timing parameters (e.g., offset, cycle length, maximum green) across all intersections, aiming to minimize arterial travel time while constraining waiting time and queue length on minor roads, using TGDT as a surrogate deep-learning model within a Bayesian optimization \cite{jones1998efficient} framework.

\section{Acknowledgments}
The work was supported in part by NSF CNS 1922782. The opinions, findings, and conclusions expressed in this publication are those of the authors and not necessarily those of NSF.


\bibliographystyle{abbrv}

\bibliography{ref}

\end{document}